\documentclass[runningheads]{llncs}
\usepackage{graphicx}
\usepackage{hyperref}

\begin{document}

\title{Bird Species Classification using Transfer Learning with Multistage Training}
%

%
%
\author{Sourya Dipta Das*\inst{1}\and
Akash Kumar*\inst{2}}
\authorrunning{Sourya Dipta Das* , Akash Kumar*}
%
\institute{Jadavpur University, Kolkata , India , \email{dipta.juetce@gmail.com} \and
Delhi Technological University, Delhi , India \\
\email{kmr2907akash@gmail.com}}
\maketitle    
%
\begin{abstract}
Bird species classification has received more and more attention in the field of computer vision, for its promising applications in biology and environmental studies. Recognizing bird species is difficult due to the challenges of discriminative region localization and fine-grained feature learning. In this paper, we have introduced a Transfer learning based method with multistage training. We have used both Pre-Trained Mask-RCNN and a ensemble model consists of Inception Nets( Inceptionv3 net \& InceptionResnetv2 ) to get the both localization and species of the bird from the images. Our final model achieves an F1 score of 0.5567 or 55.67\% on the dataset provided in CVIP 2018 Challenge. 

\keywords{Bird species classification \and Deep Networks \and Transfer Learning \and Multistage Training \and Object Detection}

\end{abstract}

\footnote{Code is available at:
\href{https://github.com/AKASH2907/bird-species-classification}{https://github.com/AKASH2907/bird-species-classification}}

\section{Introduction}
Bird species are recognized as useful biodiversity indicators. They are responsive to changes in sensitive ecosystems, whilst populations-level changes in behavior are both visible and quantifiable. Suffered from great species variation, it is difficult for non-professionals to identify the sub-category of a bird only by its appearance. However, it is exhausting to annotate all the images by human beings with expert knowledge. Thus, an automatic classification system for bird species are needed, which will be great convenience for many practical applications.
For researchers working outdoors, shoot photos can be classified and analyzed immediately by the system, illustration books are no more needed. For the public, the system could provide much fun when combined with culture information like poems and legends. It will arouse people’s interest in birds and could benefit the protections of birds. Apart from that, Classifying bird species is an interesting problem for Fine-grained categorization, also known as subcategory recognition, which is a subfield in object recognition. In recent years, fine-grained classification stood out from basic-level classification, bringing promising applications and new challenges to computer vision society.

In this Bird Species Classification Challenge, our main focus was to classify birds from high resolution photographs taken from camera. In this task, to improve classification task, we have also provided localization of birds in the respective images with their class labels. The Main Challenges involving in this problem are given below.

\begin{enumerate}

\item large Intensity variation in images as pictures are taken in different time of a day (like morning, noon, evening etc.) 
\item various poses of Bird (like flying, sitting with different orientation)
\item bird localization in the image as there are some images in which there are more than one bird in that image
\item Large Variation in Background of the images
\item various type of occlusions of birds in the images due to leaf or branches of the tree
\item Size or portion of the bird covered in the images
\item less no of sample images per class and also class imbalance

\end{enumerate}

We have proposed a end to end deep learning based approach using transfer learning to learn both micro and macro level features from Bird ROIs. We have used pre-trained Mask-RCNN to get the Bird ROIs from the images and used Multistage training method to remove class imbalance partially \& boost up the accuracy of our model. Further information about model is given in the respective section.

\section{Related Work}
In recent years, there are number of existing works to automate the classification of bird using audio data rather images. For this Purpose, Feature Extraction from audio signals has some advantages like species have distinctive calls and no line of sight is need for detection. However, there are some disadvantages also like an individual bird may emit no audio at all for extended period of time and can’t able to count the no of birds precisely. Due to this reasons, there are growing number of studies to use computer vision and image-based techniques for this problem ~\cite{ref_proc2} ~\cite{ref_proc4} ~\cite{ref_jr1}. Atanbori et al~\cite{ref_jr1} proposed a method to use motion features including curvature and wing beat frequency. Combined with Normal Bayes classifier and a Support Vector Machine classifier. Cheng et al~\cite{ref_proc2} introduced discriminative features for bird species classification based on parts of birds and Marini et al~\cite{ref_proc4}  proposed an approach to use a color segmentation to eliminate background elements and compute normalized color histograms to extract feature vector for classification. 
Bird Species Classification with visual data is also important in domain of fine grained classification and there are significant contribution to bird species classification problem respect to this domain~\cite{ref_proc1}~\cite{ref_proc3}~\cite{ref_proc6}~\cite{ref_proc7}. There are also some few shot and deep learning based approaches~\cite{ref_proc1}~\cite{ref_proc6}~\cite{ref_proc17}   regarding this problem which have achieved considered amount of accuracy in their respective datasets.

\section{Dataset}
In this paper, we have used the dataset from the CVIP 2018 Bird Species challenge. Training dataset consists of 150 images with 16 species of birds and testing dataset contains 158 images. The dataset contains high resolution images ranging from 800 $ \times $ 600 to 4000 $\times$ 6000. The dataset is class imbalanced with 5 images in one species to 20 images in other species.

\section{Proposed Approach}

\subsection{Data Augmentation}
 To increase the number of training samples per class and reduce the affect of class imbalance, data augmentation is used. Relevant image augmentation techniques are chosen according bird type of each classes. Those techniques are Gaussian Noise, Gaussian Blur, Flip,	Contrast, Hue, Add (add some values to each channel of the pixel), Multiply (multiply some values to each channel of the pixel), Sharp, Affine transform. As the dataset was quite small, the networks trained on the dataset, overfitted the dataset and does not generalize well on 150 images. After data augmentation, training dataset increased from 150 images to 1330 images.
 
\begin{table}[]
\centering
\caption{Table for Data Augmentation techniques for each of bird species}
\begin{tabular}{|l|l|l|l|l|l|l|l|l|l|l|}
\hline
Species & Gaussian Noise & Gaussian Blur & Flip & Contrast & Hue & Add & Multiply & Sharp & Affine & Total \\ \hline
blasti & Yes & Yes & Yes & Yes & Yes & No& No& No& No& 90 \\ \hline
bonegl & Yes & Yes & Yes & Yes & Yes & Yes & Yes & Yes & Yes & 78 \\ \hline
brhkyt & Yes & Yes & Yes & Yes & Yes & Yes & Yes & Yes & Yes & 65 \\ \hline
cbrtsh & Yes & Yes & Yes & Yes & Yes & Yes & Yes & Yes & Yes & 91 \\ \hline
cmnmyn & Yes & Yes & Yes & Yes & Yes & Yes & Yes & Yes & Yes & 91 \\ \hline
gretit & Yes & Yes & Yes & Yes & Yes & Yes & Yes & Yes & Yes & 78 \\ \hline
hilpig & Yes & Yes & Yes & Yes & Yes & Yes & Yes & No& No& 80 \\ \hline
himbul & Yes & Yes & Yes & Yes & Yes & No& No& No& No& 99 \\ \hline
himgri & Yes & Yes & Yes & Yes & Yes & No& No& No& No& 100 \\ \hline
hsparo & Yes & Yes & Yes & Yes & Yes & No& No& No& No& 81 \\ \hline
indvul & Yes & Yes & Yes & Yes & Yes & No& No& No& No& 81 \\ \hline
jglowl & Yes & Yes & Yes & Yes & Yes & Yes & Yes & Yes & Yes & 78 \\ \hline
lbicrw & Yes & Yes & Yes & Yes & Yes & Yes & Yes & Yes & Yes & 78 \\ \hline
mgprob & Yes & Yes & Yes & Yes & Yes & Yes & Yes & Yes & Yes & 78 \\ \hline
rebimg & Yes & Yes & Yes & Yes & Yes & Yes & Yes & No& No& 80 \\ \hline
wcrsrt & Yes & Yes & Yes & Yes & Yes & Yes & Yes & No& No& 80 \\ \hline
\end{tabular}

\label{my-label}
\end{table}

\subsection{Bird ROI(Region of Interest) Detection}
To eliminate background elements or regions and also extract features from only body of the birds, Pretrained Object Detection deep nets are used. In this Model, we have used Mask R-CNN~\cite{ref_proc8} to localize birds in each image from both test \& training dataset.We have used the pretrained weights of Mask R-CNN,trained on the COCO dataset~\cite{ref_proc18} which contains 1.5 million object instances with 80 object categories(including birds). 
 
\subsection{Transfer Learning}
Here, we have used transfer learning based approach to learn both micro and macro level feature extracted from bird images for classification.we have used ImageNet~\cite{ref_proc10} pretrained weights to initialize our Deepnet model for training. ImageNet contains 1.2 million images belonging to 1000 classes.Training using pretrained ImageNet weights help us to learn fine-grained as well as global level features before hand and learn the deepnet more specific \& discriminative features for each bird species which leads to increase the accuracy of our model.

\begin{figure}
\includegraphics[width=\textwidth]{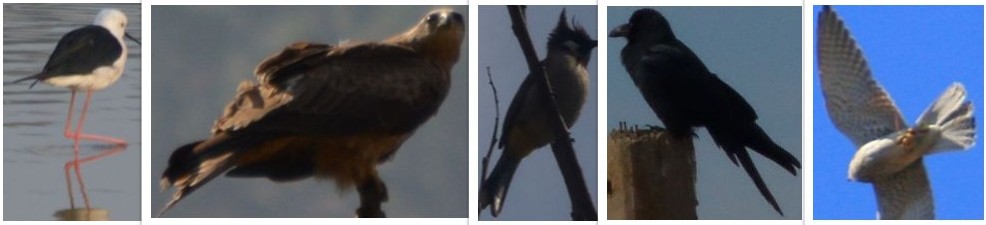}
\caption{Mask R-CNN cropped birds images} \label{fig1}
\end{figure}

\subsection{Ensemble Model Architecture}
We have used InceptionResNetV2~\cite{ref_proc12} \& InceptionV3~\cite{ref_proc14} deepnet architectures to create a ensemble model as our classification model. 
The prediction vector from Inception V3~\cite{ref_proc14} and Inception ResNet V2~\cite{ref_proc12} weights are generated for each image at the time of testing. There are two cases with Mask R-CNN: 

\hfill \break
1) Birds Detection: If the Mask R-CNN detect birds in the image, then a batch of cropped bird images are created. The whole batch for that particular image is evaluated using both network weights. Both the prediction vectors are compared and then the specie is assigned based on the prediction value with the highest weight or prediction confidence value of the specie of bird is finally predicted.

\hfill \break
2) No Bird Detection: Whole Image is predicted for the specie of bird using both the architecture weights. Though, the number of such cases are very less. The specie with the highest predicted value is added to final prediction vector.

\begin{figure}
\includegraphics[width=\textwidth]{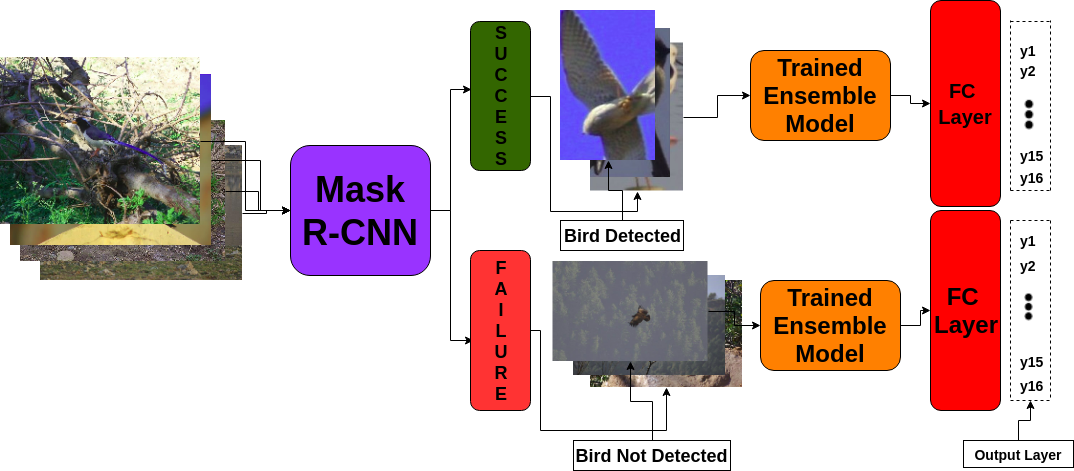}
\caption{Overall Architecture of our bird species classification system} \label{fig2}
\end{figure}

Fig. \ref{fig2} illustrates the overall process of bird detection using Mask R-CNN and species classification using ImageNet models.

\section{Experiments}
\subsection{Multi-Stage Training }
We used multi-stage training to improve the accuracy of the model. Firstly, we trained the Inception V3 architecture and then Inception ResNet V2 architecture on data augmented original images.  In the second stage, we used the pretrained weight on original images to train on cropped images generated from Mask R-CNN. All the images are resized to 416 $\times $416. The accuracy of the model increased by 2-3 \% after training on the cropped images. The multi-stage training helps to learn fine-grained features using cropped images of birds and the original images are used to learn the global spatial features present in the image. 


\subsection{Testing}
At the time of testing, the images are passed from Mask R-CNN pre-loaded with COCO weights. The cases of Mask R-CNN is discussed in the Ensemble Model Architecture. Here, we have used 'categorical cross-entropy' as loss function \&
Adam as a Optimizer to train both deep networks.For fine-tuning of the deepnet models, we have tested on various types of activation functions and Swish activation function~\cite{ref_proc15} performed best among all of them.

\section{Evaluation}
\subsubsection{Evaluation Metric}
For the challenge, we used three evaluation metrics - 
\begin{enumerate}
    \item Precision: It is the ratio of correctly predicted positive observations to the total observations. It is defined as:
    \begin{equation}
        \frac{True Positives}{True Positives + False Positives}
    \end{equation}
    
    \item Recall: It is the ratio of correctly predicted positive observations to all the observations in the relevant class.
    \begin{equation}
        \frac{True Positives}{True Positives + False Negatives}
    \end{equation}
    \item F1-score: It is the harmonic mean of precision and recall. 
    \begin{equation}
        2*\frac{precision*recall}{precision + recall}
    \end{equation}
\end{enumerate}

The evaluation metrics are calculated from Confusion matrix. In confusion matrix, True positives is equal to sum of diagonal elements. False positives is equal to the sum of each column excluding diagonal elements. False negatives is equal to the sum of each row elements excluding diagonal elements.

\section{Results}
Table-\ref{tab2} contains the F1-scores obtained from different architectures trained on original images and Mask R-CNN crops as discussed in Multi-Stage Training.
\hfill \break The Inception ResNet V2 model trained firstly on resized original images and then on Mask R-CNN cropped images gives the best results at the time of training. The final trained weights of this model was used to predict the specie of bird.
\begin{table}
\centering
\caption{Accuracy during Multi-Stage Training on InceptionResnetV2 \& InceptionV3 Models.}\label{tab2}
\begin{tabular}{|l|l|l|l|l|}
\hline
\textbf{Model Architecture} &  \textbf{Data Subset} & \textbf{Train} & \textbf{Validation} & \textbf{Test}\\
\hline
Inception V3 &  Images & 91.26 &12.76&30.95\\
\cline{2-5}
 &  Images + Crops & 93.97&15.50&41.66\\
 \cline{1-5}
Inception Resnet V2 & Images & 97.29& 29.17&47.96\\
 \cline{2-5}
 & Images + Crops & 92.29 &33.69&49.09\\

\hline
\end{tabular}
\end{table}
\hfill  \break
Table-\ref{tab3} contains the class averaged precision, recall and F1 scores. In ensemble method, the prediction vector for both Inception V3 and Inception Resnet V2 is compared for each class predicted and then it was appended to the final prediction file.
\begin{table}
\centering
\caption{Evaluation Metrics (in \%) on Test Dataset}\label{tab3}
\begin{tabular}{|l|l|l|l|}
\hline
\textbf{Model Architecture} &  \textbf{Precision} & \textbf{Recall} & \textbf{F1}\\
\hline
Mask R-CNN + InceptionV3 & 48.61&45.65 & 47.09 \\
\hline
Mask R-CNN + InceptionResnetV2 & 53.62&48.72 & 51.05\\
\hline
Mask R-CNN + Ensemble Model &  \textbf{56.58} & \textbf{54.8}&\textbf{55.67}\\
\hline

\end{tabular}
\end{table}

\begin{figure}
\includegraphics[width=\textwidth,height=8cm]{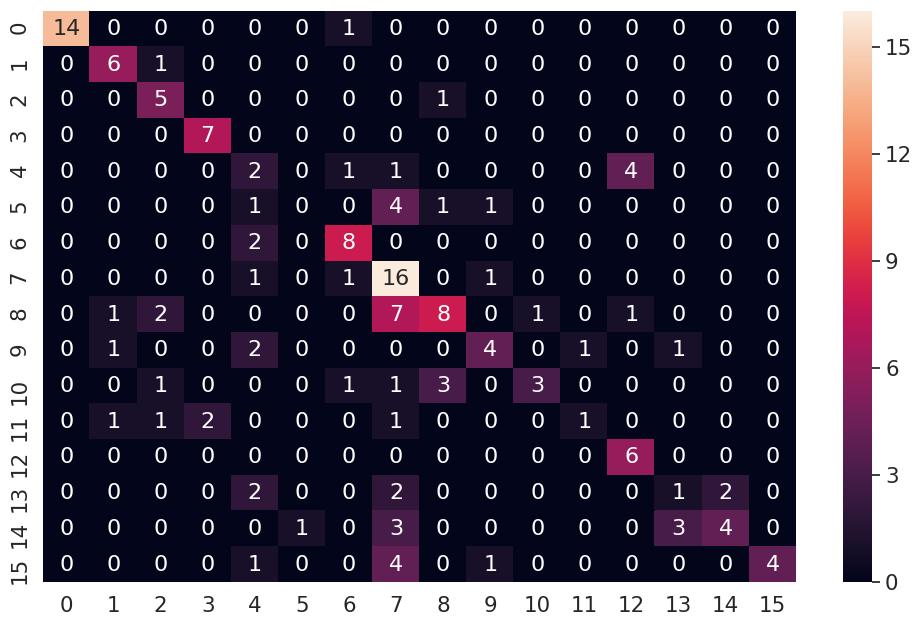}
\caption{Confusion Matrix of Mask R-CNN + Ensemble Model } \label{fig3}
\end{figure}

\section{Drawbacks}
From the confusion matrix, cmnmyn, gretit, hsparo, indvul, mgprob and rebimg classes have poor accuracy.Few training samples per class with large variance and class imbalance are the Main Reason for those misclassifications. Though there are some other reasons like small bird ROIs , similarity in bird body part's colour \& background colour and lighting condition change in training and testing dataset. For those conditions deep nets can't be able to learn those discriminative features ( both micro features like colour, gradients,textures etc. and macro features like shape , colour patch etc.). Various lighting condition ( like picture taken during daylight ,dawn, dusk, evening etc.) affected our model most as due to low background light, many micro features of the bird like colour,texture , gradients etc. are lost. Different poses of bird also reduced our model's accuracy though it is compensated with micro level features of the bird. Possible solution for those problems are discussed further in the later section.
\begin{figure}
\includegraphics[width=\textwidth]{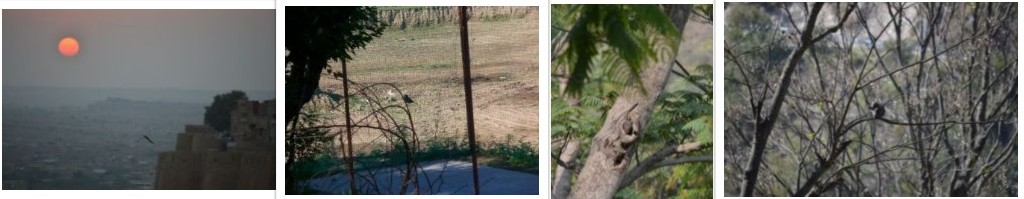}
\caption{Few example of images where Mask R-CNN fails to detect birds} \label{fig4}
\end{figure}

\section{Conclusion \& Future Work}

In this paper, we have proposed a method to both localize and classify the specie of the bird from high-definition photographs taken from camera by using a end-to-end approach with Mask R-CNN, transfer learning and multi-stage training. We have considered this challenge and as a few shot classification problem and proposed to use bird localization in images to boost up the accuracy of our model. Transfer learning helped our model to learn more specific and converge loss function more quickly with good accuracy on test dataset. In future, we are planning to extend this work using Part Model based approach with NTM (Neural Turing Machine) and Visual Attention Network.

%
%
%
%

\end{document}